\documentclass{article}

\usepackage{microtype}
\usepackage{graphicx}
\usepackage{subcaption}
\usepackage{booktabs} 
\usepackage{makecell}

\usepackage{hyperref}

\usepackage[preprint]{icml2026}

\usepackage{amsmath}
\usepackage{amssymb}
\usepackage{mathtools}
\usepackage{amsthm}

\usepackage[capitalize,noabbrev]{cleveref}

\theoremstyle{plain}

\theoremstyle{definition}

\theoremstyle{remark}

\usepackage[textsize=tiny]{todonotes}

\icmltitlerunning{Submission and Formatting Instructions for ICML 2026}

\begin{document}

\twocolumn[
  \icmltitle{Value-and-Structure Alignment for Routing-Consistent Quantization of Mixture-of-Experts Models}



  \icmlsetsymbol{equal}{*}

  \begin{icmlauthorlist}
    \icmlauthor{Hancheol Park}{nota}
    \icmlauthor{Geonho Lee}{nota}
    \icmlauthor{Tairen Piao}{nota}
    \icmlauthor{Tae-Ho Kim}{nota}
  \end{icmlauthorlist}

  \icmlaffiliation{nota}{Nota Inc., South Korea}

  \icmlcorrespondingauthor{Hancheol Park}{hancheol.park@nota.ai}
  \icmlcorrespondingauthor{Tae-Ho Kim}{thkim@nota.ai}

  \icmlkeywords{Machine Learning, ICML}

  \vskip 0.3in
]



\printAffiliationsAndNotice{}  

\begin{abstract}
Mixture-of-Experts (MoE) models scale foundation models efficiently by activating only a subset of experts for each token, but their large number of  expert parameters still makes quantization essential for practical deployment. Unlike dense models, however, MoE models are sensitive to routing instability: small quantization-induced perturbations can change the top-$k$ expert selection, altering the computation path and degrading model quality. We propose Value-and-Structure Routing Alignment for Quantization (VSRAQ), a MoE-specific post-training quantization objective that preserves pre-quantization expert-selection behavior under quantization. VSRAQ combines two complementary objectives that jointly preserve expert-selection behavior: value alignment, which matches routing-relevant logits or scores, and structure alignment, which preserves expert ordering and top-$k$ decision boundaries. By maintaining routing consistency, VSRAQ reduces quantization-induced degradation without introducing any inference-time overhead and can be integrated into existing quantization frameworks. Experiments on recent MoE foundation models show that VSRAQ improves expert-selection consistency and consistently outperforms reconstruction-only and router-aware baselines.
\end{abstract}

\section{Introduction}

Mixture-of-Experts (MoE) architectures have emerged as an effective approach for scaling large language models by increasing model capacity while activating only a subset of parameters for each input token~\citep{shazeer2017outrageously,fedus2022switch,chen2025eacmoe}. In a typical MoE layer, a router computes expert scores from the input hidden states and selects the top-$k$ experts to process each token.
This sparse activation makes MoE models attractive for efficient inference, since the number of activated parameters can remain relatively small even when the total number of model parameters is large.

However, MoE models still present a significant deployment challenge. Although only a small number of experts are activated for each token, the full set of expert parameters must generally be stored in memory. As a result, MoE models can impose substantial memory and bandwidth costs during inference.
Post-training quantization (PTQ) is therefore an important technique for deploying MoE models under practical memory and latency constraints. Existing PTQ methods for large language models, such as GPTQ~\citep{frantar2022gptq}, SmoothQuant~\citep{xiao2023smoothquant}, AWQ~\citep{lin2024awq}, and learning-based rounding methods~\citep{cheng2023optimize}, have shown strong performance on dense transformer models.

Despite their success, conventional quantization objectives are not fully suited to MoE architectures, because they are primarily designed to preserve numerical fidelity in a fixed computation graph. In contrast, MoE layers rely on input-dependent routing, in which a router selects a small subset of experts for each token. As a result, the behavior of a quantized MoE model depends not only on how accurately weights, activations, or block outputs are reconstructed, but also on whether the same experts are selected after quantization. Even small perturbations in hidden states or router logits can change the selected top-$k$ experts. Such rerouting does not merely introduce a continuous numerical error; it sends tokens through different experts and therefore changes the computation path itself. This makes MoE quantization especially sensitive to router perturbations and motivates quantization objectives that explicitly preserve routing behavior.

Recent studies have begun to address MoE-specific quantization challenges.
Some methods focus on constructing better calibration sets for MoE layers by balancing expert activation or modeling token-expert affinity~\citep{hu2025moequant, fu2025eaquant}.
Others study structure-aware mixed-precision quantization, where different MoE components are assigned different bit-widths according to their sensitivity~\citep{li2024quantmoe}.
The most closely related line of work aims to preserve routing behavior before and after quantization.
For example, EAC-MoE calibrates routers to mitigate change in expert selection caused by low-bit quantization~\citep{chen2025eacmoe}, while EAQuant introduces router consistency alignment by matching router logits or routing probability distributions~\citep{fu2025eaquant}.

These prior works demonstrate that routing behavior is critical for robust MoE quantization. However, existing router-aware quantization objectives primarily preserve routing values by matching the pre- and post-quantization logits, probabilities, or scores of the originally selected top-$k$ experts.
The top-$k$ routing decision is determined not only by the numerical values of expert scores but also by their structural relationships.
For instance, the relative ordering among selected experts affects the mixture composition, and the margin between the $k$-th selected expert and nearby non-selected experts determines how stable the selection boundary is under perturbation.
A quantized model may match router scores reasonably well while still causing rank inversions or boundary violations that change expert selection.

To address this issue, we propose \emph{Value-Structure Router-Aligned Quantization (VSRAQ)}, a MoE-specific quantization objective that aligns router outputs before and after quantization.
VSRAQ preserves routing behavior through two complementary components.
First, \emph{value alignment} matches routing-relevant logits or scores, with additional emphasis on sensitive regions of the routing function.
Second, \emph{structure alignment} preserves expert ordering and the decision boundary between selected and nearby non-selected experts.
VSRAQ can be integrated into existing PTQ frameworks as an additional calibration objective and introduces no inference-time overhead.

Our contributions are summarized as follows:
\begin{itemize}
    \item We propose VSRAQ, a MoE-specific quantization objective that jointly preserves routing values and structural relationships, including sensitivity-aware value alignment, selected-expert ordering, and top-$k$ boundary margins.

    \item We show that VSRAQ is a plug-in calibration objective for existing PTQ pipelines and introduces no inference-time overhead.

    \item We validate VSRAQ on recent MoE foundation models, demonstrating improved expert-selection consistency and reduced quantization-induced performance degradation.
\end{itemize}

\section{Related Work}

\paragraph{MoE-specific quantization.}
Recent studies have explored quantization methods tailored to the sparse and input-dependent computation of MoE models.
Because expert usage is often highly imbalanced, calibration data must provide sufficient coverage across experts to avoid quantization degradation.
MoEQuant~\citep{hu2025moequant} constructs expert-balanced calibration sets and incorporates token-expert affinity into quantization.
EAQuant~\citep{fu2025eaquant} also improves sparse expert calibration by balancing calibration data at the expert level.
Beyond calibration, QuantMoE-Bench~\citep{li2024quantmoe} shows that different MoE components, such as attention layers, shared experts, routed experts, and MoE blocks, exhibit different sensitivity to quantization, motivating component-wise or expert-wise mixed precision.
These works show that MoE quantization should account for sparse expert activation, calibration imbalance, and component-level sensitivity.

\paragraph{Router-focused MoE quantization.}
A closely related line of work focuses on the router, which determines the sparse computation path of an MoE model.
Since quantization-induced perturbations can change the selected top-$k$ experts, router-aware methods optimize quantization parameters with auxiliary losses that preserve the routing behavior of the full-precision model.
EAQuant~\citep{fu2025eaquant} aligns router logits and expert-selection probabilities before and after quantization, but its full-dimensional alignment may allocate optimization effort to low-ranked experts that rarely affect top-$k$ decisions.
EAC-MoE~\citep{chen2025eacmoe} instead focuses calibration on the top-$n k$ experts to mitigate changes in expert selection after low-bit quantization.
These methods demonstrate the importance of preserving router outputs during MoE quantization.

VSRAQ further refines router alignment by jointly preserving routing values and routing structure.
First, it applies sensitivity-aware value alignment to emphasize regions where small logit perturbations can cause larger routing-score changes.
Second, it preserves structural relationships among router outputs, including selected-expert ordering and margins between selected and nearby non-selected experts.
This provides a more targeted objective for routing-stable MoE quantization.

\section{Value-and-Structure Routing Alignment for Quantization}
\label{sec:method}

We propose Value-and-Structure Routing Alignment for Quantization (VSRAQ), a MoE-specific quantization objective that preserves routing behavior during post-training quantization.
VSRAQ is not a standalone quantization algorithm, but a calibration objective that can be integrated into existing quantization frameworks.
In this work, we instantiate VSRAQ on top of a learning-based post-training quantization framework, where quantization parameters such as clipping ranges, scale factors, zero points, and rounding parameters are optimized using a small calibration dataset.

\subsection{Reconstruction-Based Quantization}

Let $F$ denote an unquantized decoder block and $\hat{F}$ its quantized counterpart.
Given an input hidden state $\mathbf{x}$, we denote the corresponding block outputs as
\begin{equation}
    \mathbf{y}^{\mathrm{fp}} = F(\mathbf{x}),
    \qquad
    \mathbf{y}^{q} = \hat{F}(\mathbf{x}).
\end{equation}
Conventional reconstruction-based quantization optimizes quantization parameters by minimizing the block output error:
\begin{equation}
    \mathcal{L}_{\mathrm{recon}}
    =
    \left\|
    \mathbf{y}^{q} - \mathbf{y}^{\mathrm{fp}}
    \right\|_2^2 .
\end{equation}

While this objective is effective for dense models, it does not explicitly preserve the routing behavior of MoE layers.
In MoE models, the output of a decoder block is strongly influenced by the experts selected by the router.
Therefore, minimizing only the reconstruction loss can lead to solutions where the block outputs are numerically close, but the selected experts are changed after quantization.

To address this, VSRAQ augments the reconstruction objective with a router alignment loss:
\begin{equation}
    \mathcal{L}
    =
    \mathcal{L}_{\mathrm{recon}}
    +
    \lambda_{\mathrm{router}}
    \mathcal{L}_{\mathrm{router}},
    \label{eq:total_loss}
\end{equation}
where $\lambda_{\mathrm{router}}$ controls the strength of router alignment.

\subsection{Router Logits}

Let $\mathbf{z}^{\mathrm{fp}} \in \mathbb{R}^{E}$ denote the router logits from the unquantized model, where $E$ is the number of experts.
Similarly, let $\mathbf{z}^{q} \in \mathbb{R}^{E}$ denote the router logits from the quantized model.
We sort the experts according to $\mathbf{z}^{\mathrm{fp}}$ and denote by $\pi(i)$ the expert index ranked $i$-th under the unquantized router logits.
That is, $\pi$ is a permutation of expert indices satisfying
\begin{equation}
    z^{\mathrm{fp}}_{\pi(1)}
    \geq
    z^{\mathrm{fp}}_{\pi(2)}
    \geq
    \cdots
    \geq
    z^{\mathrm{fp}}_{\pi(E)} .
\end{equation}
The top-$k$ selected expert set is then defined as
\begin{equation}
    \mathcal{T}_k
    =
    \{\pi(1), \pi(2), \ldots, \pi(k)\}.
\end{equation}

VSRAQ decomposes router alignment into value alignment and structure alignment:
\begin{equation}
    \mathcal{L}_{\mathrm{router}}
    =
    \mathcal{L}_{\mathrm{value}}
    +
    \mathcal{L}_{\mathrm{struct}} .
    \label{eq:router_loss}
\end{equation}

\subsection{Value Alignment}

The value alignment loss preserves routing-relevant logits or scores before and after quantization.
VSRAQ decomposes value alignment into two components: direct top-$k$ logit preservation and sigmoid-aware score preservation.

\paragraph{Top-$k$ logit preservation.}
A natural objective is to directly match the quantized router logits to those of the original full-precision model for the experts selected by the unquantized router.
Following EAC-MoE~\citep{chen2025eacmoe}, we define the TopK-MSE loss as
\begin{equation}
    \mathcal{L}_{\mathrm{TopK}}
    =
    \frac{1}{k}
    \sum_{i=1}^{k}
    \left(
    z^{q}_{\pi(i)}
    -
    z^{\mathrm{fp}}_{\pi(i)}
    \right)^2 .
    \label{eq:topk_mse}
\end{equation}

\paragraph{Sigmoid-aware score preservation.}
Direct logit matching does not consider the routing form used by recent MoE models.
Modern MoE architectures such as DeepSeek-V3~\citep{deepseekai2024deepseekv3},
Nemotron-3-Nano~\citep{blakeman2025nemotron3nano}, and
GLM-4.5~\citep{zeng2025glm45} adopt sigmoid-based routing.
Unlike softmax routing, which normalizes scores across all experts and induces direct competition among them,
sigmoid routing estimates token--expert affinities independently before top-$k$ selection,
often normalizing only the selected experts.
Due to the S-shaped curve of the sigmoid function, scores near $0.5$ change much more sharply with respect to input logit perturbations than saturated scores near $0$ or $1$.
As a result, quantization errors can cause larger changes in routing scores near this region and may alter expert selection.
VSRAQ addresses this issue by preserving sigmoid-based routing scores before and after quantization.

To account for this, VSRAQ introduces a sigmoid-aware value alignment loss.
For each selected expert $\pi(i)$, we compute
\begin{equation}
    p^{\mathrm{fp}}_{\pi(i)}
    =
    \sigma
    \left(
    z^{\mathrm{fp}}_{\pi(i)}
    \right),
    \qquad
    p^{q}_{\pi(i)}
    =
    \sigma
    \left(
    z^{q}_{\pi(i)}
    \right),
\end{equation}
where $\sigma(\cdot)$ denotes the sigmoid function.
We define a sensitivity weight as
\begin{equation}
    w_{\pi(i)}
    =
    \left(
    p^{\mathrm{fp}}_{\pi(i)}
    \left(
    1 - p^{\mathrm{fp}}_{\pi(i)}
    \right)
    \right)^{\gamma}
    +
    \epsilon_w ,
    \label{eq:sigmoid_weight}
\end{equation}
where $\gamma$ controls the sharpness of the weighting and $\epsilon_w$ is a small minimum weight for numerical stability.
Since $p(1-p)$ is maximized at $p=0.5$, this weighting assigns larger importance to logits in the sensitive region of the sigmoid function. 

The sigmoid-aware alignment loss is defined as
\begin{equation}
    \mathcal{L}_{\mathrm{sig}}
    =
    \frac{1}{k}
    \sum_{i=1}^{k}
    w_{\pi(i)}
    \left(
    p^{q}_{\pi(i)}
    -
    p^{\mathrm{fp}}_{\pi(i)}
    \right)^2 .
    \label{eq:sigmoid_loss}
\end{equation}

The sensitivity weight $w_{\pi(i)}$ prevents this loss from over-constraining
the quantization optimization. VSRAQ optimizes quantization parameters under
multiple objectives to preserve pre-quantization routing decisions. Therefore,
forcing the model to precisely match saturated scores, which have limited
influence on the routing decision, can consume calibration capacity and
interfere with more routing-critical objectives. The proposed weight suppresses such low-sensitivity score discrepancies and focuses the calibration signal on regions where score changes are more likely to affect routing.

The final value alignment loss is
\begin{equation}
    \mathcal{L}_{\mathrm{value}}
    =
    (1 - \alpha)
    \mathcal{L}_{\mathrm{TopK}}
    +
    \alpha
    \mathcal{L}_{\mathrm{sig}},
    \label{eq:value_loss}
\end{equation}
where $\alpha \in [0,1]$ controls the balance between direct logit matching and sigmoid-aware score matching.
In practice, we set $\alpha$ to a relatively small value, as preserving the logits of the originally selected top-$k$ experts is the main driver of routing consistency, while the sigmoid-aware term serves as an auxiliary refinement for sensitive routing regions.

\subsection{Structure Alignment}

Value alignment encourages the top-$k$ router logits or scores to remain close before and after quantization.
However, routing stability depends not only on the values themselves, but also on their structural relationships.
In particular, VSRAQ preserves two routing-relevant structures: the ordering among selected experts and the margin around the top-$k$ selection boundary.

\paragraph{Top-$k$ order preservation.}
The selected experts are sorted by the full-precision router as
\begin{equation}
    z^{\mathrm{fp}}_{\pi(1)}
    \geq
    z^{\mathrm{fp}}_{\pi(2)}
    \geq
    \cdots
    \geq
    z^{\mathrm{fp}}_{\pi(k)} .
\end{equation}
VSRAQ encourages the quantized router to preserve this ordering.
For adjacent selected experts, we define the rank preservation loss as
\begin{equation}
    \mathcal{L}_{\mathrm{rank}}
    =
    \frac{1}{k-1}
    \sum_{i=1}^{k-1}
    \mathrm{softplus}
    \left(
    m_{\mathrm{rank}}
    -
    \left(
    z^{q}_{\pi(i)}
    -
    z^{q}_{\pi(i+1)}
    \right)
    \right),
    \label{eq:rank_loss}
\end{equation}
where $m_{\mathrm{rank}}$ is a rank margin.
When $m_{\mathrm{rank}}=0$, this loss penalizes rank inversions.
When $m_{\mathrm{rank}}>0$, it additionally encourages a positive margin between adjacent selected experts.

\paragraph{Boundary preservation.}
Rerouting often occurs near the top-$k$ cutoff boundary, especially between the $k$-th selected expert and nearby non-selected experts.
Let $b$ denote the number of nearby non-selected experts considered after the cutoff.
For $j=1,\ldots,b$, we define the full-precision and quantized boundary margins as
\begin{equation}
    \Delta^{\mathrm{fp}}_j
    =
    z^{\mathrm{fp}}_{\pi(k)}
    -
    z^{\mathrm{fp}}_{\pi(k+j)},
    \qquad
    \Delta^{q}_j
    =
    z^{q}_{\pi(k)}
    -
    z^{q}_{\pi(k+j)} .
\end{equation}
The boundary margin preservation loss is
\begin{equation}
    \mathcal{L}_{\mathrm{margin}}
    =
    \frac{1}{b}
    \sum_{j=1}^{b}
    \left(
    \Delta^{q}_j
    -
    \Delta^{\mathrm{fp}}_j
    \right)^2 .
    \label{eq:margin_loss}
\end{equation}

The final structure alignment loss is
\begin{equation}
    \mathcal{L}_{\mathrm{struct}}
    =
    \beta_{\mathrm{rank}}
    \mathcal{L}_{\mathrm{rank}}
    +
    \beta_{\mathrm{margin}}
    \mathcal{L}_{\mathrm{margin}},
    \label{eq:struct_loss}
\end{equation}
where $\beta_{\mathrm{rank}}$ and $\beta_{\mathrm{margin}}$ control the relative strengths of order preservation and boundary preservation.

\subsection{Overall Objective}

Combining the value and structure terms, the final router alignment loss is
\begin{equation}
    \mathcal{L}_{\mathrm{router}}
    =
    \mathcal{L}_{\mathrm{value}}
    +
    \mathcal{L}_{\mathrm{struct}}.
\end{equation}
The total calibration objective is then given by Eq.~\ref{eq:total_loss}.
Since VSRAQ modifies only the calibration objective, it does not change the inference graph of the quantized model and introduces no additional inference-time overhead.

\section{Experiments}
\label{sec:experiments}

\subsection{Experimental Setup}

We evaluate VSRAQ on two recent open MoE LLMs: Solar-Open-100B~\citep{park2026solaropen} and Nemotron-3-Nano-30B-A3B~\citep{blakeman2025nemotron3nano}.
Solar-Open-100B is a 102B-parameter MoE language model, and we use it as a large-scale Transformer-based MoE evaluation target.
Nemotron-3-Nano-30B-A3B is included because it adopts a hybrid Mamba-Transformer MoE architecture, allowing us to evaluate whether VSRAQ remains effective beyond standard Transformer-only MoE models. For Solar-Open-100B, we evaluate W4A16 weight-only quantization and NVFP4 quantization. While W4A16 quantizes only the model weights and keeps activations in higher precision, NVFP4 quantizes both weights and activations. For Nemotron-3-Nano-30B-A3B, we evaluate NVFP4 quantization.

We use AutoRound~\citep{cheng2023optimize} as the base post-training quantization framework.
AutoRound optimizes quantization parameters by minimizing a block-wise reconstruction loss on a small calibration dataset.
Our baseline corresponds to AutoRound with its original reconstruction objective.
VSRAQ is implemented by augmenting this reconstruction objective with the proposed router alignment loss, so that the quantization parameters are optimized to preserve both block outputs and routing behavior. We also compare against a variant that uses the TopK-MSE loss from EAC-MoE~\citep{chen2025eacmoe} as an auxiliary loss in AutoRound.

For calibration, we use a compact reasoning-oriented calibration set, denoted
as the NVIDIA Reasoning calibration set, constructed from
OpenCodeReasoning~\citep{nvidia_opencodereasoning},
OpenScienceReasoning-2~\citep{nvidia_opensciencereasoning2},
and OpenMathReasoning~\citep{nvidia_openmathreasoning}.
For each model, we randomly sample 512 examples from this mixture. We use a
maximum calibration sequence length of 2,048 tokens for Solar-Open-100B and
1,024 tokens for Nemotron-3-Nano-30B-A3B.

We evaluate quantization performance from three perspectives using
lm-evaluation-harness v0.4.11 \cite{eval-harness}.
First, we report perplexity on WikiText-2~\citep{merity2016pointer}.
Second, we evaluate reasoning-oriented generation performance with
generation-based answer extraction, where the model generates a free-form response
and the final answer is extracted from the generated text. This category includes
MMLU-Pro~\citep{wang2024mmlu_pro}, GPQA-Diamond~\citep{rein2023gpqa}, and
GSM-8K~\citep{cobbe2021gsm8k}. For MMLU-Pro and GPQA-Diamond, we use a thinking
budget of 8,192 tokens. Third, we evaluate general short-form understanding performance with
likelihood-based multiple-choice evaluation, where the model scores each candidate
answer by log-likelihood and selects the highest-scoring option. This category
includes ARC-C, ARC-E~\citep{clark2018think}, BoolQ~\citep{clark2019boolq},
HellaSwag~\citep{zellers2019hellaswag}, MMLU~\citep{hendrycks2021mmlu},
PIQA~\citep{bisk2020piqa}, TruthfulQA~\citep{lin2022truthfulqa}, and
WinoGrande~\citep{sakaguchi2020winogrande}.

For each benchmark category, we report the average score across the corresponding
tasks. For all generation-based answer extraction evaluations, we use greedy
decoding rather than sampling. Our goal is not to maximize benchmark scores through
decoding-time tuning, but to measure the performance degradation introduced by
quantization relative to the original model. Greedy decoding removes
sampling-induced randomness and allows us to evaluate quantization effects in a
more reproducible manner.

\begin{table}[t]
  \caption{VSRAQ hyperparameter configuration.}
  \label{tab:vsraq_hyperparameters}
  \begin{center}
    \begin{small}
      \begin{tabular}{lc}
        \toprule
        Hyperparameter & Value \\
        \midrule
        $b$ & $1$ \\
        $\alpha$ & $0.3$ \\
        $\gamma$ & $1.0$ \\
        $\epsilon_w$ & $10^{-3}$ \\
        $m_{\mathrm{rank}}$ & $0.0$ \\
        $\beta_{\mathrm{rank}}$ & $0.01$ \\
        $\beta_{\mathrm{margin}}$ & $0.05$ \\
        $\lambda_{\mathrm{router}}$ & $\{0.1, 0.01\}$ \\
        \bottomrule
      \end{tabular}
    \end{small}
  \end{center}
  \vskip -0.1in
\end{table}

We set the VSRAQ hyperparameters so that the router alignment loss has a
comparable scale to the reconstruction loss and produces gradients of a similar magnitude during calibration. The same hyperparameter configuration is used across all models and quantization settings, except for the router loss coefficient $\lambda_{\mathrm{router}}$. Specifically, we use
$\lambda_{\mathrm{router}}=0.01$ for Solar-Open-100B
and $\lambda_{\mathrm{router}}=0.1$ for Nemotron-3-Nano-30B-A3B. The configuration is summarized in Table~\ref{tab:vsraq_hyperparameters}. VSRAQ is applied only
during calibration and does not add any additional computation during inference. Consequently, except for differences in generation quality,
VSRAQ has the same memory footprint and inference latency as the corresponding
AutoRound-quantized model at the same precision.

\begin{table}[t]
\centering
\begin{minipage}{\columnwidth}
\centering
\caption{Benchmark results for Solar-Open-100B under W4A16 quantization.
PPL denotes perplexity. Gen.-based Extract. denotes generation-based answer
extraction, and Likelihood-based MC denotes likelihood-based multiple-choice
evaluation. Lower PPL is better, and higher benchmark scores are better. Bold indicates the best result among quantized methods.}
\label{tab:solar_w4a16}
\footnotesize
\setlength{\tabcolsep}{3.5pt}
\begin{tabular}{lccc}
\toprule
\makecell[c]{Method}
& \makecell[c]{PPL}
& \makecell[c]{Gen.-based\\Extract.}
& \makecell[c]{Likelihood-\\based MC} \\
\midrule
Unquantized & 6.06 & 73.62 & 76.92 \\
\specialrule{\lightrulewidth}{0pt}{0pt}
AutoRound   & 7.12 & 72.19 & 76.44 \\
TopK-MSE    & 6.87 & 71.66 & 76.72 \\
VSRAQ       & \textbf{6.81} & \textbf{73.22} & \textbf{76.76} \\
\bottomrule
\end{tabular}
\end{minipage}
\end{table}

\begin{table}[t]
\centering
\begin{minipage}{\columnwidth}
\centering
\caption{Benchmark results for Solar-Open-100B under NVFP4 quantization.}
\label{tab:solar_nvfp4}
\footnotesize
\setlength{\tabcolsep}{3.5pt}
\begin{tabular}{lccc}
\toprule
\makecell[c]{Method}
& \makecell[c]{PPL}
& \makecell[c]{Gen.-based\\Extract.}
& \makecell[c]{Likelihood-\\based MC} \\
\midrule
Unquantized & 6.06 & 73.62 & 76.92 \\
\specialrule{\lightrulewidth}{0pt}{0pt}
AutoRound   & 7.22 & 61.78 & 75.59 \\
TopK-MSE    & 7.10 & 63.86 & 75.21 \\
VSRAQ       & \textbf{6.90} & \textbf{64.43} & \textbf{75.63} \\
\bottomrule
\end{tabular}
\end{minipage}
\end{table}

\begin{table}[t]
\centering
\begin{minipage}{\columnwidth}
\centering
\caption{Benchmark results for Nemotron-3-Nano-30B-A3B under NVFP4 quantization.}
\label{tab:nemotron_nvfp4}
\footnotesize
\setlength{\tabcolsep}{3.5pt}
\begin{tabular}{lccc}
\toprule
\makecell[c]{Method}
& \makecell[c]{PPL}
& \makecell[c]{Gen.-based\\Extract.}
& \makecell[c]{Likelihood-\\based MC} \\
\midrule
Unquantized & 10.36 & 73.64 & 70.33 \\
\specialrule{\lightrulewidth}{0pt}{0pt}
AutoRound   & 10.79 & 69.10 & 69.96 \\
TopK-MSE    & 10.78 & 69.47 & 69.59 \\
VSRAQ       & \textbf{10.70} & \textbf{70.92} & \textbf{70.18} \\
\bottomrule
\end{tabular}
\end{minipage}
\end{table}

\subsection{Results}
\label{sec:results}

Tables~\ref{tab:solar_w4a16}, \ref{tab:solar_nvfp4}, and
\ref{tab:nemotron_nvfp4} summarize our experimental results. VSRAQ consistently
achieves the best performance among the quantized methods across all settings.
For likelihood-based multiple-choice evaluation, 4-bit quantization introduces
only a small performance drop compared with the unquantized model. Consequently,
the absolute improvement over baselines is modest, but VSRAQ still obtains the
highest average score, showing that the proposed routing-aware objective remains
beneficial even when the evaluation is relatively insensitive to small routing
differences.

In contrast, the improvements are much more visible on generation-based answer
extraction tasks. These tasks require the model to produce free-form reasoning
before the final answer is extracted, making them more sensitive to errors that
accumulate during decoding. In MoE models, routing decisions at each generation
step influence which experts process the hidden states, and the resulting hidden
states in turn affect subsequent token predictions. Therefore, small
expert-selection mismatches introduced by quantization can compound over long
generations. VSRAQ mitigates this effect by explicitly preserving routing
behavior during calibration, leading to clearer gains over AutoRound and the
TopK-MSE variant.

\section{Discussion}
\label{sec:discussion}

\subsection{Effect of Each VSRAQ Component}

We first analyze whether each component of VSRAQ contributes to quantization quality.
Table~\ref{tab:ablation} reports PPL on WikiText-2 as components are added step by step.

\begin{table}[t]
  \caption{Step-by-step ablation study on WikiText-2 PPL. ``+ Sigmoid'' adds sigmoid-aware value alignment on top of TopK-MSE, and ``+ Structure'' further adds structure alignment. Lower is better.}
  \label{tab:ablation}
  \begin{center}
    \begin{small}
      \begin{tabular}{lccc}
        \toprule
        Method & \shortstack{Solar\\W4A16} & \shortstack{Solar\\NVFP4} & \shortstack{Nemotron\\NVFP4} \\
        \midrule
        Unquantized   & 6.06 & 6.06 & 10.36 \\
        \specialrule{\lightrulewidth}{0pt}{0pt}
        AutoRound     & 7.12 & 7.22 & 10.79 \\
        + TopK-MSE & 6.87 & 7.10 & 10.78 \\
        + Sigmoid     & 6.72 & 6.98 & 10.79 \\
        + Structure   & \textbf{6.81} & \textbf{6.90} & \textbf{10.70} \\
        \bottomrule
      \end{tabular}
    \end{small}
  \end{center}
  \vskip -0.1in
\end{table}

TopK-MSE loss consistently improves over the baseline, showing that preserving routing-relevant logits is useful for MoE quantization.
Adding sigmoid-aware alignment further improves PPL for Solar-Open-100B, suggesting that sensitivity-aware weighting can better preserve routing scores in sigmoid-based routing.
The full VSRAQ objective, which additionally includes structure alignment, gives the best PPL for the NVFP4 settings and remains substantially better than the baseline in all cases.
Although the effect of each component is not strictly monotonic in every setting, the overall trend suggests that both value alignment and structure alignment contribute to more robust MoE quantization.

\subsection{Expert-Selection Consistency}

To directly measure whether VSRAQ preserves routing decisions, we compute the agreement between the top-$k$ experts selected by the full-precision and quantized models.
For each token and each MoE layer, we compute the Jaccard similarity between the two selected expert sets:
\begin{equation}
    \mathrm{Jaccard}
    \left(
    \mathcal{T}^{\mathrm{fp}}_k,
    \mathcal{T}^{q}_k
    \right)
    =
    \frac{
    \left|
    \mathcal{T}^{\mathrm{fp}}_k
    \cap
    \mathcal{T}^{q}_k
    \right|
    }{
    \left|
    \mathcal{T}^{\mathrm{fp}}_k
    \cup
    \mathcal{T}^{q}_k
    \right|
    } .
\end{equation}

We use 616,759 tokens from the stem, chat, math, and code subsets of Nemotron-Post-Training-Dataset-v2.\footnote{\url{https://huggingface.co/datasets/nvidia/Nemotron-Post-Training-Dataset-v2}}
VSRAQ consistently improves the agreement between pre- and post-quantization expert selection across layers.
At the largest margin, VSRAQ achieves more than an 11.29\% relative improvement over the baseline in expert-selection agreement.
This result supports the central motivation of VSRAQ: explicitly preserving router values and routing structure leads to more stable expert selection after quantization.

\begin{figure}[t]
  \vskip 0.2in
  \begin{center}
    \centerline{\includegraphics[width=0.95\columnwidth]{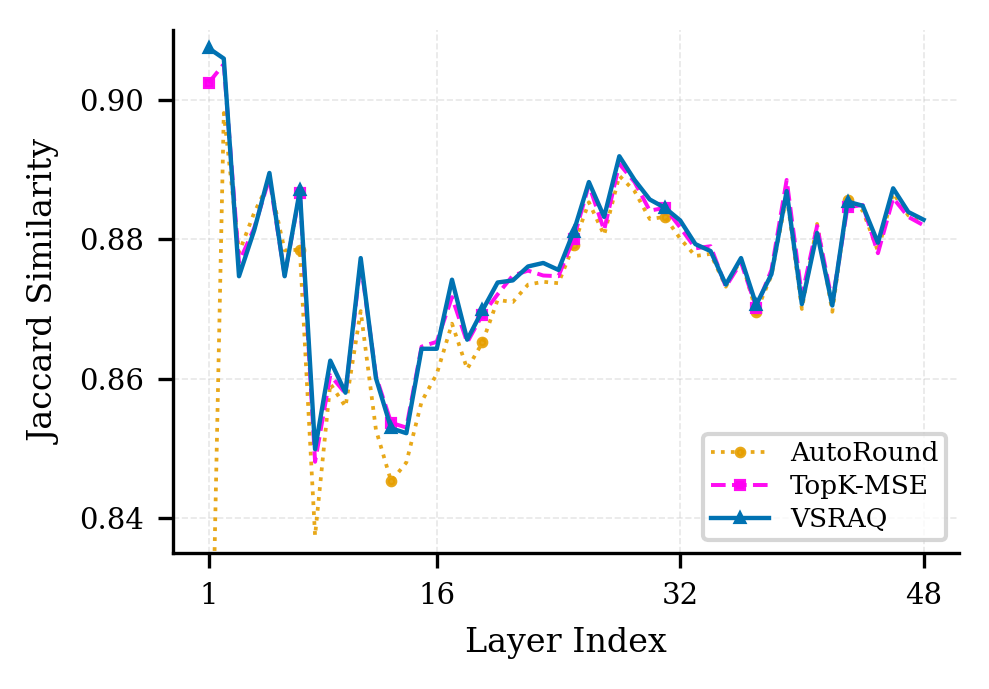}}
    \caption{
      Layer-wise Jaccard Similarity between the top-$k$ expert sets selected by the full-precision and quantized models under W4A16 quantization.
VSRAQ better preserves expert-selection consistency than AutoRound and TopK-MSE.
    }
    \label{fig:jaccard}
  \end{center}
  \vskip -0.1in
\end{figure}

\section{Conclusion}

We presented Value-and-Structure Routing Alignment for Quantization (VSRAQ), a MoE-specific
post-training quantization objective that jointly preserves routing values and
routing structure. VSRAQ improves perplexity, downstream benchmark performance,
and expert-selection consistency over reconstruction-only and TopK-MSE baselines,
with clearer gains on generation-based evaluations where routing mismatches can
accumulate during decoding.

Our experiments focus on 4-bit settings, including W4A16 and NVFP4 quantization,
where routing-aware calibration may have a modest effect due to relatively limited
quantization noise. We expect larger benefits in more aggressive 2-bit or 3-bit
regimes, where expert-selection instability is likely to be more severe. We leave
this extension, along with broader evaluations across MoE architectures and
quantization frameworks, to future work.

\section*{Impact Statement}

This paper presents work whose goal is to advance efficient deployment of
mixture-of-experts (MoE) language models through MoE-specific post-training
quantization. By reducing the performance degradation caused by low-bit
quantization, VSRAQ may help make large MoE models easier to deploy with lower
memory and computational cost. This can contribute to improved accessibility and
energy efficiency. However, as with other quantization methods for large language
models, further investigation is needed to understand whether quantization-induced
changes in model outputs may affect undesirable behaviors such as unethical text
generation, biased outputs, or hallucinations. Practical deployment of quantized
MoE models should therefore be accompanied by appropriate safety, robustness, and
fairness evaluations.


\bibliography{example_paper}
\bibliographystyle{icml2026}

\end{document}